\documentclass[letterpaper, 10 pt, conference]{ieeeconf}  

\IEEEoverridecommandlockouts                              

\usepackage{amsmath} 
\usepackage{amsfonts}
\usepackage[table,xcdraw]{xcolor}
\usepackage{multirow}
\usepackage{float}
\usepackage{wrapfig}
\usepackage{graphicx}
\usepackage{hyperref}

\title{\LARGE \bf
A Robust and Efficient Visual-Inertial Initialization with  Probabilistic Normal Epipolar Constraint}

\author{Changshi Mu$^{1}$, Daquan Feng$^{1\dagger}$, Qi Zheng$^{1\dagger}$, and Yuan Zhuang$^{2}$
\thanks{$^{1}$Changshi Mu, Daquan Feng, and Qi Zheng are with the Guangdong Key Laboratory of Intelligent Information Processing, College of Electronics and Information Engineering, Shenzhen University, Shenzhen 518060, China
        {\tt\small \{fdquan, qiz\}@szu.edu.cn, 2200432055@email.szu.edu.cn}}%
\thanks{$^{2}$Yuan Zhuang is with the State Key Laboratory of Information Engineering in Surveying, Mapping and Remote Sensing, Wuhan University, Wuhan 430072, China {\tt\small \{yuan.zhuang\}@whu.edu.cn}}
\thanks{$^{\dagger}$Corresponding authors
        }%
}

\begin{document}

\maketitle
\thispagestyle{empty}
\pagestyle{empty}

\begin{abstract}

Accurate and robust initialization is essential for Visual-Inertial Odometry (VIO), as poor initialization can severely degrade pose accuracy. During initialization, it is crucial to estimate parameters such as accelerometer bias, gyroscope bias, initial velocity, gravity, etc. Most existing VIO initialization methods adopt Structure from Motion (SfM) to solve for gyroscope bias. However, SfM is not stable and efficient enough in fast-motion or degenerate scenes. To overcome these limitations, we extended the rotation-translation-decoupled framework by adding new uncertainty parameters and optimization modules. First, we adopt a gyroscope bias estimator that incorporates probabilistic normal epipolar constraints. Second, we fuse IMU and visual measurements to solve for velocity, gravity, and scale efficiently. Finally, we design an additional refinement module that effectively reduces gravity and scale errors. Extensive EuRoC dataset tests show that our method reduces gyroscope bias and rotation errors by 16\% and 4\% on average, and gravity error by 29\% on average. On the TUM dataset, our method reduces the gravity error and scale error by 14.2\% and 5.7\% on average respectively. The source code is available at \href{https://github.com/MUCS714/DRT-PNEC.git}{https://github.com/MUCS714/DRT-PNEC.git}.


\end{abstract}


\section{INTRODUCTION}

Visual-Inertial Odometry (VIO) aims to estimate camera position in unknown environments by fusing camera images and IMU measurements. The camera estimates a visual map and reduces pose drift, while the IMU provides a metric scale for motion and short-term robustness. VIO has many advantages, such as small size, low cost, and low power consumption, leading to increasing applications in virtual reality \cite{fang2017real}, augmented reality \cite{li2017monocular,piao2017adaptive}, and automated robotics \cite{sun2018robust,yang2017real}.

To effectively run a VIO system, parameters such as scale, gravity direction, initial velocity, and sensor biases must be accurately estimated during initialization. Incorrect initialization leads to poor convergence and inaccurate parameter estimation. In addition, fast initialization is important since the VIO system cannot function without proper IMU initialization \cite{campos2020inertial}.

Basically, previous VIO initialization works are tightly or loosely coupled. Tightly coupled methods \cite{martinelli2014closed}, \cite{kaiser2016simultaneous}, \cite{dominguez2018visual} approximate camera poses from IMU, fuse visual and IMU data, and use closed-form solutions, increasing cost and often ignoring gyroscope bias, harming accuracy. Loosely coupled methods \cite{campos2020inertial,qin2017robust,mur2017visual} assume accurate visual SfM-derived trajectories, solve SfM first, and initialize inertial parameters based on camera poses, relying heavily on SfM performance, which can be unstable in fast motion or with few common feature points.

Overall, both tightly coupled and loosely coupled methods fail to fully exploit the complementary information between the camera and the IMU. Specifically, tightly coupled methods do not utilize visual observations to estimate gyroscope bias, which can lead to numerical stability issues and lower accuracy. Loosely coupled methods do not use IMU measurements to enhance the stability of visual SfM, resulting in low accuracy or failure of initialization in challenging motion scenarios. Inspired by the fact that image observations can be directly used to optimize the rotation between camera frames \cite{kneip2013direct}, He et al. \cite{he2023rotation} proposed a rotation-translation-decoupled VIO initialization method. It first estimates the gyroscope bias through the gyroscope bias estimator, and then estimates the rotation and translation independently. This method enhances the connection between visual observations and IMU measurements. Wang et al. \cite{wang2024stereo} extended this framework to the stereo visual-inertial SLAM system and improved translation estimation through 3-DoF bundle adjustment, which significantly promoted the performance of the SLAM system. However, the gyroscope bias estimator overlooks the quality of image feature matches, thus giving each match an equal weight in the final result. Even though outliers are removed from feature matches, error distributions of 2D feature correspondences vary with image content and the specific matching technique. Therefore, it is crucial to consider the uncertainty of 2D feature matches.

To overcome the limitations of SfM and improve the accuracy and robustness of initialization, we propose a new initialization method based on the rotation-translation-decoupling framework \cite{he2023rotation}. This method increases the accuracy of the gyroscope bias estimation and reduces errors in the scale and gravity directions.
In summary, the contributions of this work include:
\begin{itemize}
    \item We propose a gyroscope bias estimator with the Probabilistic Normal Epipolar Constraint (PNEC). Based on the 3D covariance of unit bearing vector and IMU pre-integration, we reconstruct the variance of the Normal Epipolar Constraint (NEC) residual distribution and successfully introduce this variance into the gyroscope bias estimator.
    \item We incorporate a modified scale-gravity refinement module, which effectively refines only scale and gravity without considering other parameters.
    \item We compare our method with other initialization counterparts. The experimental results demonstrate that our method achieves more accurate gyroscope bias estimation and lower average errors.
\end{itemize}

\section{related work}
Initialization in VIO systems is critical because it affects the accuracy and robustness of the systems. Numerous initialization methods have been proposed and applied to VIO systems (e.g., \cite{campos2021orb,qin2018vins,mur2017visual,geneva2020openvins}) in recent years. Martinelli \cite{martinelli2014closed} proposed for the first time a tightly coupled closed-form solution to jointly recover parameters including initial velocity, gravity, and feature point depth. His method assumes that common feature points can be observed in all frames during the initialization process, and IMU measurements can be used to estimate camera pose. However, this method is unsuitable for inexpensive and noisy IMU sensors as it ignores gyroscope bias. 
Subsequently, Kaiser et al. extended this work in \cite{kaiser2016simultaneous}. They iteratively solved a nonlinear least-squares problem that includes the gravity magnitude to determine the gyroscope bias. Their experiments demonstrated that gyroscope bias affects the accuracy of closed-form solutions. Nevertheless, the tightly coupled closed-form solution suffers from low accuracy and computational efficiency in estimating gyroscope bias.

With the emergence of higher precision visual odometry or SfM \cite{engel2017direct,mur2017orb}, the loosely coupled method using precise camera poses to solve IMU initialization parameters has been proposed \cite{mur2017visual,qin2017robust}. Mur-Artal and Tardós \cite{mur2017visual} process the IMU and visual initialization separately. They calculate the initial estimation of scale, gravity, velocity, and IMU biases based on a set of keyframe poses processed by the monocular SLAM algorithm. Similarly, Qin and Shen \cite{qin2017robust} proposed a linear system but set the accelerometer bias to zero in visual-inertial bundle adjustment. Both methods ignore the uncertainty of sensors and the correlation between inertial parameters. To solve this problem, Campos et al. \cite{campos2020inertial} proposed a maximum-a-posteriori framework to initialize IMU parameters. Zu\~niga-No\"el et al. \cite{zuniga2021analytical} proposed a non-iterative analytical solution for estimating IMU parameters within a maximum-a-posteriori framework.

\section{PRELIMINARIES}

\subsection{Visual-Inertial Notation} In this paper, we define the notation as follows. The IMU frame and the camera frame at the time index $i$ are represented by $\mathbf{F}_{b_i}$ and $\mathbf{F}_{c_i}$, respectively. Let $\mathbf{R}_{b_{i}b_{j}}$ and $\mathbf{p}_{b_{i}b_{j}}$ denote the rotation and translation between the IMU frame at time index $i$ and the IMU frame at time index $j$. Define the gravity vector as $\mathbf{g}=(0,0,G)^{\top}$, where $G$ is the magnitude of gravity. The camera and IMU are rigidly attached, and the transformation $\mathbf{T}_{bc} = [\mathbf{R}_{bc} | \mathbf{p}_{bc}]$ between their reference systems is determined by calibration. $\left\lfloor\cdot\right\rfloor_{\times}$ and $\left \| \cdot \right \|$ denote the skew-symmetric operation and the Euclidean norm operation. 
At two time points corresponding to IMU frames $\mathbf{F}_{b_i}$ and $\mathbf{F}_{b_j}$, we pre-integrate linear acceleration and angular velocity within the local frame $\mathbf{F}_{b_i}$. Let $\boldsymbol{\alpha}_{b_{j}}^{b_{i}}$, $\boldsymbol{\beta}_{b_{j}}^{b_{i}}$, $\boldsymbol{\gamma}_{b_{j}}^{b_{i}}$ represent the pre-integration of translation, velocity, and rotation from $\mathbf{F}_{b_i}$ to $\mathbf{F}_{b_j}$:


\begin{equation}
\boldsymbol{\alpha}_{b_{j}}^{b_{i}}=\sum_{k=i}^{j-1}\left(\left(\sum_{f=i}^{k-1} \mathbf{R}_{b_{i} b_{f}} \mathbf{a}_{f}^{m} \Delta t\right) \Delta t+\frac{1}{2} \mathbf{R}_{b_{i} b_{k}} \mathbf{a}_{k}^{m} \Delta t^{2}\right) 
\end{equation}
\begin{equation}
\boldsymbol{\beta}_{b_{j}}^{b_{i}}=\sum_{k=i}^{j-1} \mathbf{R}_{b_{i} b_{k}} \mathbf{a}_{k}^{m} \Delta t
\end{equation}
\begin{equation}
\boldsymbol{\gamma}_{b_{j}}^{b_{i}}=\prod_{k=i}^{j-1} \operatorname{Exp}\left(\boldsymbol{\omega}_{k}^{m} \Delta t\right) 
\end{equation} 

\noindent where $\mathrm{Exp}(\cdot)$ stands for the exponential map $\mathrm{Exp}:\mathfrak{s o}(3)  \to SO(3)$. $\omega^m_k$ and $a^m_k$ represent the gyroscope and accelerometer measurements at time $k$ respectively, and $\Delta t$ denotes the time interval between successive IMU data. The above pre-integration formula is independent of the bias. We use the rotation pre-integration update formula in \cite{forster2016manifold}. The effect of the gyroscope bias $\mathbf{b}_g$ on the rotation pre-integration $\boldsymbol{\gamma}_{b_{j}}^{b_{i}}$ can be expressed as a first-order Taylor approximation:
\begin{equation}
\hat{\boldsymbol{\gamma}}_{b_{j}}^{b_{i}}=\boldsymbol{\gamma}_{b_{j}}^{b_{i}} \operatorname{Exp}\left(\mathbf{J}_{\mathbf{b}_{g}}^{\boldsymbol{\gamma}_{b_{j}}^{b_{i}}} \mathbf{b}_{g}\right)
\label{eq:4}
\end{equation}
where $\mathbf{J}_{\mathbf{b}_{g}}^{\boldsymbol{\gamma}_{b_{j}}^{b_{i}}}$ denotes the Jacobian of the derivative of $\boldsymbol{\gamma}_{b_{j}}^{b_{i}}$ with respect to $\mathbf{b}_g$. This Jacobian is a constant that can be efficiently computed iteratively \cite{forster2016manifold}. In this work, we ignore the accelerometer bias as in \cite{kaiser2016simultaneous} since this has little effect on the initialization result.

The motion between two consecutive keyframes can be computed by integrating the inertial measurements. We use the standard approach on $SO(3)$ manifold described in \cite{forster2016manifold}:
\begin{equation}
\mathbf{p}_{c_{0} b_{j}}=\mathbf{p}_{c_{0} b_{i}}+\mathbf{v}_{b_{i}}^{c_{0}} \Delta t_{i j}-\frac{1}{2} \mathbf{g}^{c_{0}} \Delta t_{i j}^{2}+\mathbf{R}_{c_{0} b_{i}} \boldsymbol{\alpha}_{b_{j}}^{b_{i}}
\label{eq:5}
\end{equation}
\begin{equation}
\mathbf{v}_{b_{j}}^{c_{0}}=\mathbf{v}_{b_{i}}^{c_{0}}-\mathbf{g}^{c_{0}} \Delta t_{i j}+\mathbf{R}_{c_{0} b_{i}} \boldsymbol{\beta}_{b_{j}}^{b_{i}}
\label{eq:6}
\end{equation}
\begin{equation}
\mathbf{R}_{c_{0} b_{j}}=\mathbf{R}_{c_{0} b_{i}} \boldsymbol{\gamma}_{b_{j}}^{b_{i}}
\end{equation}
where $\mathbf{R}_{c_{0} b_{j}}$ denotes the rotation from camera frame at time index 0 (i.e., the first camera frame) to the IMU frame at time index $j$. $\mathbf{p}_{c_{0} b_{j}}$ represents the corresponding translation. $\mathbf{v}_{b_{j}}^{c_{0}}$ and $\mathbf{g}^{c_{0}}$ denote the IMU velocity at time index $j$ and gravity in the $\mathbf{F}_{c_0}$ coordinate system, respectively. $\Delta t_{i j}$ is the time interval from time index $i$ to time index $j$. 

\subsection{Background – NEC}
Next, we revisit the essence of the normal epipolar constraint (NEC) from \cite{kneip2012finding}. The NEC characterizes the feature constraint between two camera frames, comprising the bearing vectors of the frames and the normal vectors of the epipolar plane. As in Fig. \ref{pic:1}, when $\mathbf{F}_{c_i}$ and $\mathbf{F}_{c_j}$ view the same 3D point $\Theta_{k}$, they form an epipolar plane with $\Theta_{k}$. Its normal vector is $\mathbf{n}_{k}=\left\lfloor\mathbf{f}_{i}^{k}\right\rfloor_{\times} \mathbf{R}_{c_{i} c_{j}} \mathbf{f}_{j}^{k}$, where $\mathbf{f}_{i}^{k}$ and $\mathbf{f}_{j}^{k}$ are unit bearing vectors from $\mathbf{F}_{c_i}$ and $\mathbf{F}_{c_j}$ to $\Theta_{k}$. All normal vectors, perpendicular to $\mathbf{p}_{c_{i} c_{j}}$, define the epipolar normal plane. Ideally, they're coplanar, enabling us to set the constraint residual on the normalized epipolar error: 

\begin{figure}[b]  
    \centering
    \includegraphics[width=0.65\columnwidth]{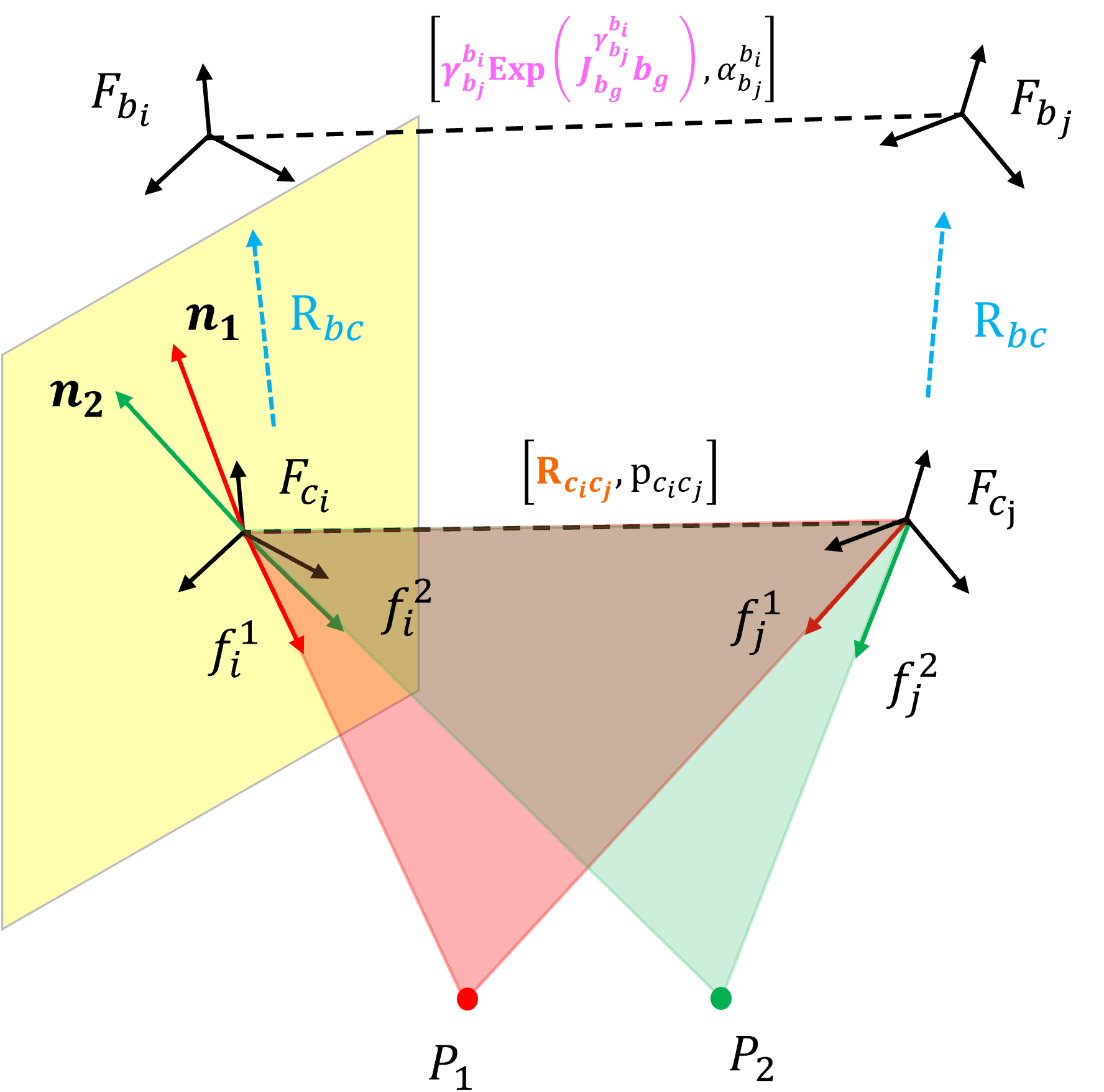}
    \caption{Geometry of the normal epipolar constraint (NEC) and the relationship between gyroscope bias and NEC. The normal vectors $\mathbf{n}_{1}$ and $\mathbf{n}_{2}$ are perpendicular to the epipolar plane where $\mathbf{f}_{i}^{1}$($\mathbf{f}_{i}^{2}$) and $\mathbf{f}_{j}^{1}$($\mathbf{f}_{j}^{2}$) are located (red and green), and all normal vectors are in the same plane (yellow), forming a constraint that can be used to solve the rotation $\mathbf{R}_{c_{i} c_{j}}$ (orange). The problem of solving $\mathbf{R}_{c_{i} c_{j}}$ is transformed into the problem of solving the gyroscope bias (pink) by using the extrinsic parameter $\mathbf{R}_{bc}$ (blue).}
    \label{pic:1}
\end{figure}

\begin{equation}
e_{k}=\left|\mathbf{p}_{c_{i} c_{j}}^{\top} \mathbf{n}_{k}\right|
\end{equation}
where $\mathbf{p}_{c_{i} c_{j}}$ denotes the translation vector from $\mathbf{F}_{c_i}$ to $\mathbf{F}_{c_j}$. The geometry of the residual is expressed as the Euclidean distance from the normal vector to the epipolar normal plane. The NEC energy function is constructed with this residual:
\begin{equation}
E(\mathbf{R}_{c_{i} c_{j}}, \mathbf{p}_{c_{i} c_{j}}) =\sum_{k}e_{k}^{2} = \sum_{k}\left|\mathbf{p}_{c_{i} c_{j}}^{\top}\left(\left\lfloor\mathbf{f}_{i}^{k}\right\rfloor_{\times} \mathbf{R}_{c_{i} c_{j}} \mathbf{f}_{j}^{k}\right)\right|^{2} 
\end{equation}

The relative rotation $\mathbf{R}_{c_{i} c_{j}}$ is estimated by ensuring the coplanarity of the normal vectors. Assuming that the two camera frames jointly observe $n$ 3D points, we can compute $n$ normal vectors of the epipolar plane and stack them into a matrix $\mathbf{N}=\left[\mathbf{n}_{1}\ldots\mathbf{n}_{n}\right]$. The requirement for coplanarity is mathematically expressed by the condition that the minimum eigenvalue of the matrix $\mathbf{M}=\mathbf{N}\mathbf{N}^{\top}$ is zero. Thus, the problem of solving the rotation can be parameterized as:
\begin{equation}
\begin{aligned}
\mathbf{R}_{c_{i} c_{j}}^{*} & =\underset{\mathbf{R}_{c_{i} c_{j}}}{\operatorname{argmin}} \lambda_{\mathbf{M}_{i j}, \min } \\
\text { with } \mathbf{M}_{i j} & =\sum^{n}\left(\left\lfloor\mathbf{f}_{i}^{k}\right\rfloor_{\times} \mathbf{R}_{c_{i} c_{j}} \mathbf{f}_{j}^{k}\right)\left(\left\lfloor\mathbf{f}_{i}^{k}\right\rfloor_{\times} \mathbf{R}_{c_{i} c_{j}} \mathbf{f}_{j}^{k}\right)^{\top}  
\end{aligned}
\label{eq:10}
\end{equation}
where $\lambda_{\mathbf{M}_{i j}, \min}$ represents the smallest eigenvalue of $\mathbf{M}_{i j}$.

Drawing inspiration from Kneip and Lynen's research \cite{kneip2013direct}, He et al.  \cite{he2023rotation} utilize the NEC approach to optimize gyroscope bias directly. This involves integrating image observations and the camera-IMU extrinsic calibration $\mathbf{T}_{bc} = [\mathbf{R}_{bc} | \mathbf{p}_{bc}]$.
\begin{equation}
\begin{aligned}
\mathbf{R}_{c_{i} c_{j}} & =\mathbf{R}_{b c}^{\top} \mathbf{R}_{b_{i} b_{j}} \mathbf{R}_{b c} \\
\mathbf{p}_{c_{i} c_{j}} & =\mathbf{R}_{b c}^{\top}\left(\mathbf{p}_{b_{i} b_{j}}+\mathbf{R}_{b_{i} b_{j}} \mathbf{p}_{b c}-\mathbf{p}_{b c}\right)
\end{aligned}
\label{eq:11}
\end{equation}

\noindent$\mathbf{R}_{b_{i} b_{j}}$ can be obtained by integrating the gyroscope measurements through Eq. \eqref{eq:4}. Substituting Eqs. \eqref{eq:11} and \eqref{eq:4} into Eq. \eqref{eq:10}, the objective function becomes:
\begin{equation}
\begin{aligned}
\mathbf{b}_{g}^{*} & =\underset{\mathbf{b}_{g}}{\operatorname{argmin}} \lambda_{\mathbf{M}_{i j}^{\prime}, \min } \\
\text { with } \mathbf{M}_{i j}^{\prime} & =\sum_{k=1}^{n}\left(\left\lfloor\mathbf{f}_{i}^{k}\right\rfloor_{\times} \mathbf{R}_{b c}^{\top} \boldsymbol{\gamma}_{b_{j}}^{b_{i}} \operatorname{Exp}\left(\mathbf{J}_{\mathbf{b}_{g}}^{\boldsymbol{\gamma}_{b_{j}}^{b_{i}}} \mathbf{b}_{g}\right) \mathbf{R}_{b c} \mathbf{f}_{j}^{k}\right) \\
 & \left(\left\lfloor\mathbf{f}_{i}^{k}\right\rfloor_{\times} \mathbf{R}_{b c}^{\top} \boldsymbol{\gamma}_{b_{j}}^{b_{i}} \operatorname{Exp}\left(\mathbf{J}_{\mathbf{b}_{g}}^{\boldsymbol{\gamma}_{b_{j}}^{b_{i}}} \mathbf{b}_{g}\right) \mathbf{R}_{b c} \mathbf{f}_{j}^{k}\right)^{\top}
\end{aligned}
\label{eq:12}
\end{equation}
which is one of the contributions in paper \cite{he2023rotation}. Fig. \ref{pic:1} illustrates the transformation in Eq. \eqref{eq:12}.

\section{proposed approach}
Accurate estimation of the gyroscope bias plays a core role in improving the trajectory accuracy of VIO systems. The bias impacts the rotation, which in turn affects the integration of both translation and velocity. In this section, we present a method that can accurately solve the initialization parameters, which include gyroscope bias, velocity, gravity, and scale. The initialization process is divided into the following four steps: (1) gyroscope bias estimation, (2) rotation and translation estimation, (3) scale, velocity, and gravity estimation, and (4) scale and gravity refinement.

\subsection{Gyroscope Bias Estimation}
Considering that image feature matches have different error distributions, Muhle et al. \cite{muhle2022probabilistic} propose the probabilistic normal epipolar constraint (PNEC). Inspired by this, we aim to estimate gyroscope bias more accurately by reducing 2D feature point position uncertainty. We introduce feature position uncertainty in the gyroscope bias estimator, assigning an anisotropic covariance matrix to each feature point. For consecutive frames $\mathbf{F}_{c_i}$ and $\mathbf{F}_{c_j}$, we apply the PNEC method to extract the 3D covariance matrix $\boldsymbol{\Sigma}_{k}$ for unit bearing vectors in $\mathbf{F}_{c_j}$. PNEC assumes 2D Gaussian position error in the image plane, with each feature having a known 2D covariance matrix $\boldsymbol{\Sigma}_{2D,k}$. Using Laplace's approximation, we can derive the 2D covariance matrix for KLT tracks from the KLT energy function. We first compute the Jacobian for each pixel $\boldsymbol{\eta}$ in the pattern $\mathcal{P}$ on $\mathbf{F}_{c_i}$. The pattern $\mathcal{P}$ is used for selecting pixels. Let $\nabla I\left(\boldsymbol{\eta}\right)$ signify the pixel gradient and $I\left(\boldsymbol{\eta}\right)$ denote the pixel intensity. Let $\mathbf{J}_{\boldsymbol{\eta}_{i}}$ denote the Jacobian with respect to the pixel position. Then, we have:
\begin{equation}
\begin{array}{c}
      \mathbf{J}_{\xi}^{i}=\left(\begin{array}{ccc}
    1 & 0 & -\boldsymbol{\eta}_{i, v} \\
    0 & 1 & \boldsymbol{\eta}_{i, u}
    \end{array}\right)\\ \\
     \mathbf{J}_{\boldsymbol{\eta}_{i}} = |\mathcal{P}| \frac{\nabla I\left(\boldsymbol{\eta}_{i}\right) \sum\limits_{\boldsymbol{\eta}_{j} \in \mathcal{P}} I\left(\boldsymbol{\eta}_{j}\right)^{\top}\mathbf{J}_{\xi}^{i} -I\left(\boldsymbol{\eta}_{i}\right) \sum\limits_{\boldsymbol{\eta}_{j} \in \mathcal{P}} \nabla I\left(\boldsymbol{\eta}_{j}\right)^{\top} \mathbf{J}_{\xi}^{j}}{\left(\sum\limits_{\boldsymbol{\eta}_{j} \in \mathcal{P}} I\left(\boldsymbol{\eta}_{j}\right)\right)^{2}}  \\
\end{array}
\end{equation}
where $\boldsymbol{\eta}_{i, u}$ and $\boldsymbol{\eta}_{i, v}$ represent the positions of pixel $\boldsymbol{\eta}_{i}$ on the image. $|\mathcal{P}|$ is the number of pixels in $\mathcal{P}$. We can obtain the covariance matrix regarding the $SE(2)$ transformation by combining all the Jacobians:
\begin{equation}
\boldsymbol{\Sigma}_{SE(2)}=\left[\left(\mathbf{J}_{\boldsymbol{\eta}_{1}}^{\top},\mathbf{J}_{\boldsymbol{\eta}_{2}}^{\top},\cdots,\mathbf{J}_{\boldsymbol{\eta}_{n}}^{\top}\right)\left(\begin{array}{c}
\mathbf{J}_{\boldsymbol{\eta}_{1}} \\
\mathbf{J}_{\boldsymbol{\eta}_{2}} \\
\vdots \\
\mathbf{J}_{\boldsymbol{\eta}_{n}}
\end{array}\right)\right]^{-1}
\end{equation}
and the upper left $2\times2$ part of $\boldsymbol{\Sigma}_{SE(2)}$ is the 2D covariance matrix of $\mathbf{F}_{c_i}$, which we define as $\boldsymbol{\Sigma}_{2D,ci}$. We then transform this matrix to $\mathbf{F}_{c_j}$ using the estimated 2D rotation $\mathbf{R}_{\theta}$:
\begin{equation}
\boldsymbol{\Sigma}_{2D,k}=\mathbf{R}_{\theta}\boldsymbol{\Sigma}_{2D,c_{i}}\mathbf{R}_{\theta}^{\top}
\end{equation}

Given the 2D covariance matrix $\boldsymbol{\Sigma}_{2D,k}$ of the feature position in $\mathbf{F}_{c_j}$, the unscented transform \cite{uhlmann1995dynamic} is applied via the unprojection function. It calculates mean and covariance by transforming selected points. First, we sample five points around each feature point, which is determined by $\boldsymbol{\mu}$ (pixel coords $[u, v]$) and $\boldsymbol{\Sigma}_{2D,k}$. Then, we apply the unscented transform as follows:
\begin{equation}
    \begin{array}{rlr}
    \boldsymbol{\xi}_{0} & =\boldsymbol{\mu} \\
    w_{0} & =\frac{1}{n+1} & \\
    \boldsymbol{\xi}_{i, i+n} & =\boldsymbol{\mu} \pm \sqrt{n+1} \boldsymbol{C}_{i} & i=1 \ldots n \\
    w_{i, i+n} & =\frac{1}{2(n+1)} & i=1 \ldots n
    \end{array}
\end{equation}
where $w$ represents the weight, $\boldsymbol{\xi}$ represents the position of the transformed point. $\boldsymbol{C}_{i}$ refers to the $i$-th column of matrix $\boldsymbol{C}$, and $\boldsymbol{C}$ is obtained from the Cholesky-decomposition of $\boldsymbol{\Sigma}_{2D,k}=\boldsymbol{C}\boldsymbol{C}^{\top}$. Then, we use the non-linear function $f(\boldsymbol{\xi})=h(g(\boldsymbol{\xi}))$ to map the points to $\mathbb{R}^3$:
\begin{equation}
\begin{aligned}
    \boldsymbol{\zeta}&=f(\boldsymbol{\xi}) \\
    g(\boldsymbol{\xi})&=K_{\text {inv }}\left(\begin{array}{c}
    \boldsymbol{\xi}_{1} \\
    \boldsymbol{\xi}_{2} \\
    1
    \end{array}\right) \\
    h(\boldsymbol{x})&=\frac{\boldsymbol{x}}{\left\|\boldsymbol{x}\right\|}
\end{aligned}
\end{equation}

The new mean and variance can be calculated:
\begin{equation}
    \begin{aligned}
    \boldsymbol{\mu}_{k} & =\sum_{i=0}^{2 n} w_{i} \boldsymbol{\zeta}_{i} \\
    \boldsymbol{\Sigma}_{k} & =\sum_{i=0}^{2 n} w_{i}\left(\boldsymbol{\zeta}_{i}-\boldsymbol{\mu}_{k}\right)\left(\boldsymbol{\zeta}_{i}-\boldsymbol{\mu}_{k}\right)^{\top}
    \end{aligned}
\end{equation}
where $\boldsymbol{\Sigma}_{k}$ is the 3D covariance matrix of the unit bearing vector $\mathbf{f}_{j}^{k}$ in $\mathbf{F}_{c_j}$. Then based on the 3D covariance, we can derive a probability distribution for the NEC residuals, which is a univariate Gaussian $\mathcal{N}\left (0,\boldsymbol{\Sigma}_{k}^{2}\right)$ with variance:
\begin{equation}
\sigma_{k}^{2}=\mathbf{p}_{ij}^{\top} \left\lfloor\mathbf{f}_{i}^{k}\right\rfloor_{\times} \mathbf{R}_{ij} \boldsymbol{\Sigma}_{k} \mathbf{R}_{ij}^{\top} {\left\lfloor\mathbf{f}_{i}^{k}\right\rfloor_{\times}}^{\top} \mathbf{p}_{ij}
\label{eq:19}
\end{equation}

The calculation of this variance requires the poses $\mathbf{R}_{ij}$ and $\mathbf{p}_{ij}$ between two frames. However, Eq. \eqref{eq:12} uses ten images to solve the gyroscope bias at one time and does not rely on the poses provided by SfM. This means that there are no variables regarding poses in the system before the estimation of the gyroscope bias. Therefore, we propose to use IMU pre-integration to reconstruct Eq. \eqref{eq:19} and provide initial poses for the variance.
\begin{equation}
\begin{aligned}
\tilde{\sigma}_{k}^{2}&=\mathbf{p}_{ij}^{\top} \left\lfloor\mathbf{f}_{i}^{k}\right\rfloor_{\times} \mathbf{R}_{ij} \boldsymbol{\Sigma}_{k} \mathbf{R}_{ij}^{\top} {\left\lfloor\mathbf{f}_{i}^{k}\right\rfloor_{\times}}^{\top} \mathbf{p}_{ij}\\
\mathbf{p}_{ij}& =\mathbf{R}_{b c}^{\top}\left(\boldsymbol{\alpha}_{b_{j}}^{b_{i}}+\boldsymbol{\gamma}_{b_{j}}^{b_{i}} \mathbf{p}_{b c}-\mathbf{p}_{b c}\right)\\
\mathbf{R}_{ij} & =\mathbf{R}_{b c}^{\top} \boldsymbol{\gamma}_{b_{j}}^{b_{i}} \mathbf{R}_{b c} \\
\end{aligned}
\end{equation}
where $\boldsymbol{\alpha}_{b_{j}}^{b_{i}}$ and $\boldsymbol{\gamma}_{b_{j}}^{b_{i}}$ denote the translation and rotation pre-integration. To integrate this variance into an eigenvalue-based gyroscope bias estimation equation, we employ an optimization scheme analogous to the well-known iteratively reweighted least-square (IRLS) algorithm \cite{lawson1961contribution}. The estimation problem is transformed into:
\begin{equation}
\begin{aligned}
\mathbf{b}_{g}^{*} & =\underset{\mathbf{b}_{g}}{\operatorname{argmin}} \lambda_{\mathbf{M}_{i j}^{\prime\prime}, \min } \\
\text { with } \mathbf{M}_{ij}^{\prime\prime}  &=\sum_{k=1}^{n}\frac{\left(\left\lfloor\mathbf{f}_{i}^{k}\right\rfloor_{\times} \mathbf{R}_{c_{i} c_{j}}\mathbf{f}_{j}^{k}\right) \left(\left\lfloor\mathbf{f}_{i}^{k}\right\rfloor_{\times} \mathbf{R}_{c_{i} c_{j}} \mathbf{f}_{j}^{k}\right)^{\top}}{\tilde{\sigma}_{k}^{2}} \\
\mathbf{R}_{c_{i} c_{j}}&=\mathbf{R}_{bc}^{\top} \boldsymbol{\gamma}_{b_{j}}^{b_{i}} \operatorname{Exp}\left(\mathbf{J}_{\mathbf{b}_{g}}^{\boldsymbol{\gamma}_{b_{j}}^{b_{i}}} \mathbf{b}_{g}\right) \mathbf{R}_{bc}
\end{aligned}
\end{equation}

So far, we have proposed a new gyroscope bias estimation formula, incorporating the 3D covariance of the unit bearing vector innovatively to reduce interference from feature point position uncertainty.

Given the gyroscope bias changes slowly during initialization, it can be assumed constant. Any keyframe pair $\left(i, j\right)\in \mathcal{E}$ with enough common features can estimate the bias. $\mathcal{E}$ represents keyframe pairs meeting initialization conditions for optimization, and the optimization problem can be simply expressed as: 
\begin{equation}
\mathbf{b}_{g}^{*}  =\underset{\mathbf{b}_{g}}{\operatorname{argmin}} \lambda 
\text { with } \lambda=\sum_{\left(i, j\right)\in \mathcal{E}}\lambda_{\mathbf{M}_{i j}^{\prime\prime}, \min }
\label{eq:22}
\end{equation}

We use the Levenberg-Marquardt algorithm with rotation parameterized by the Cayley transformation \cite{kneip2013direct} to solve Eq. \eqref{eq:22}, initializing the gyroscope bias by minimizing $\lambda$. After solving $\mathbf{b}_g$, we remove the bias and reintegrate gyroscope measurements to get accurate rotation for IMU and camera frames.

\subsection{Velocity, Gravity and Scale Estimation}
Following DRT-l, we use LiGT \cite{cai2021pose} to solve for translation. Then we use the constraints in \cite{qin2017robust} to solve for gravity, scale, and velocity.

\begin{figure*}[h] 
    \centering
    \begin{minipage}{0.9\columnwidth}
        \centering
        \includegraphics[width=\textwidth]{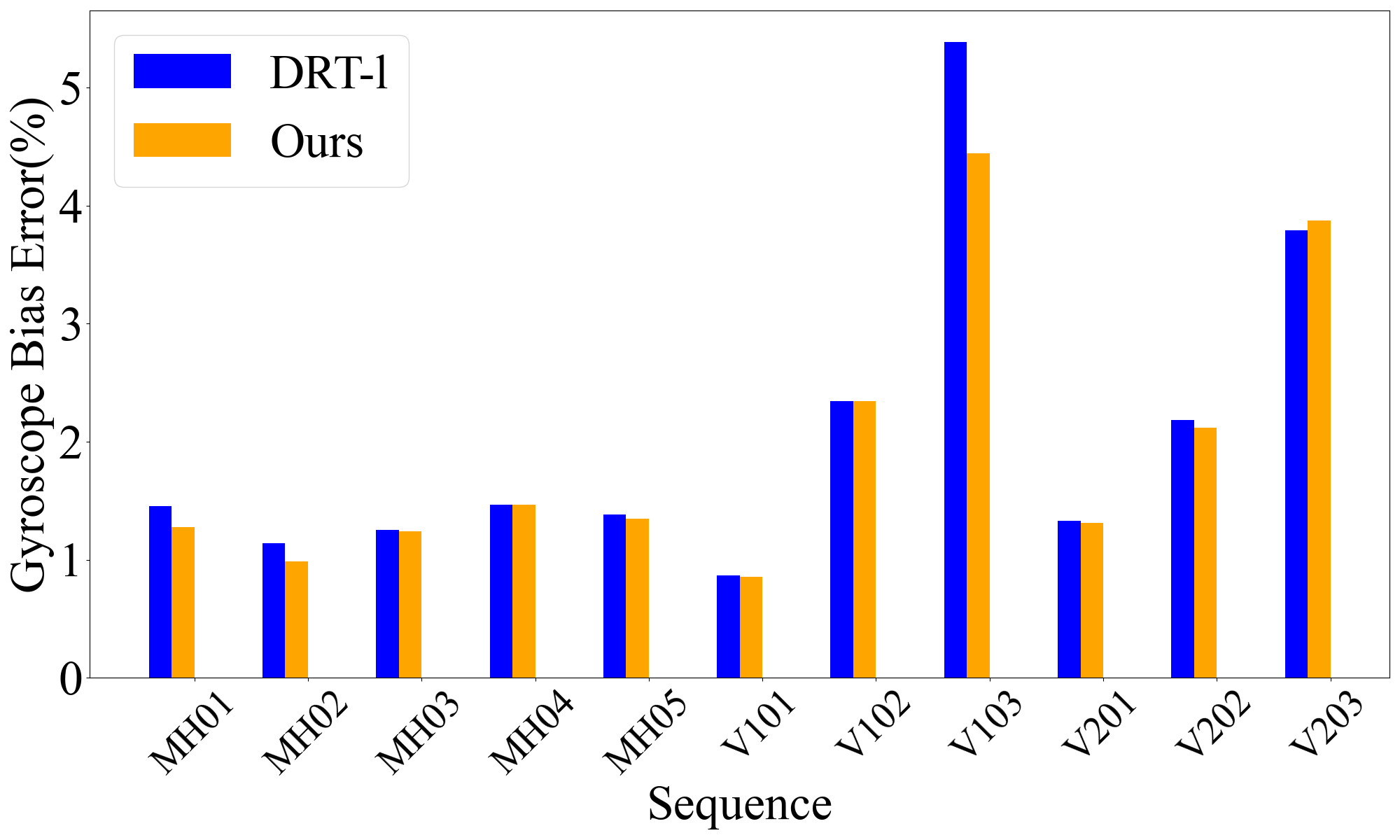}
    \end{minipage}\hfill
    \begin{minipage}{0.9\columnwidth}
        \centering
        \includegraphics[width=\textwidth]{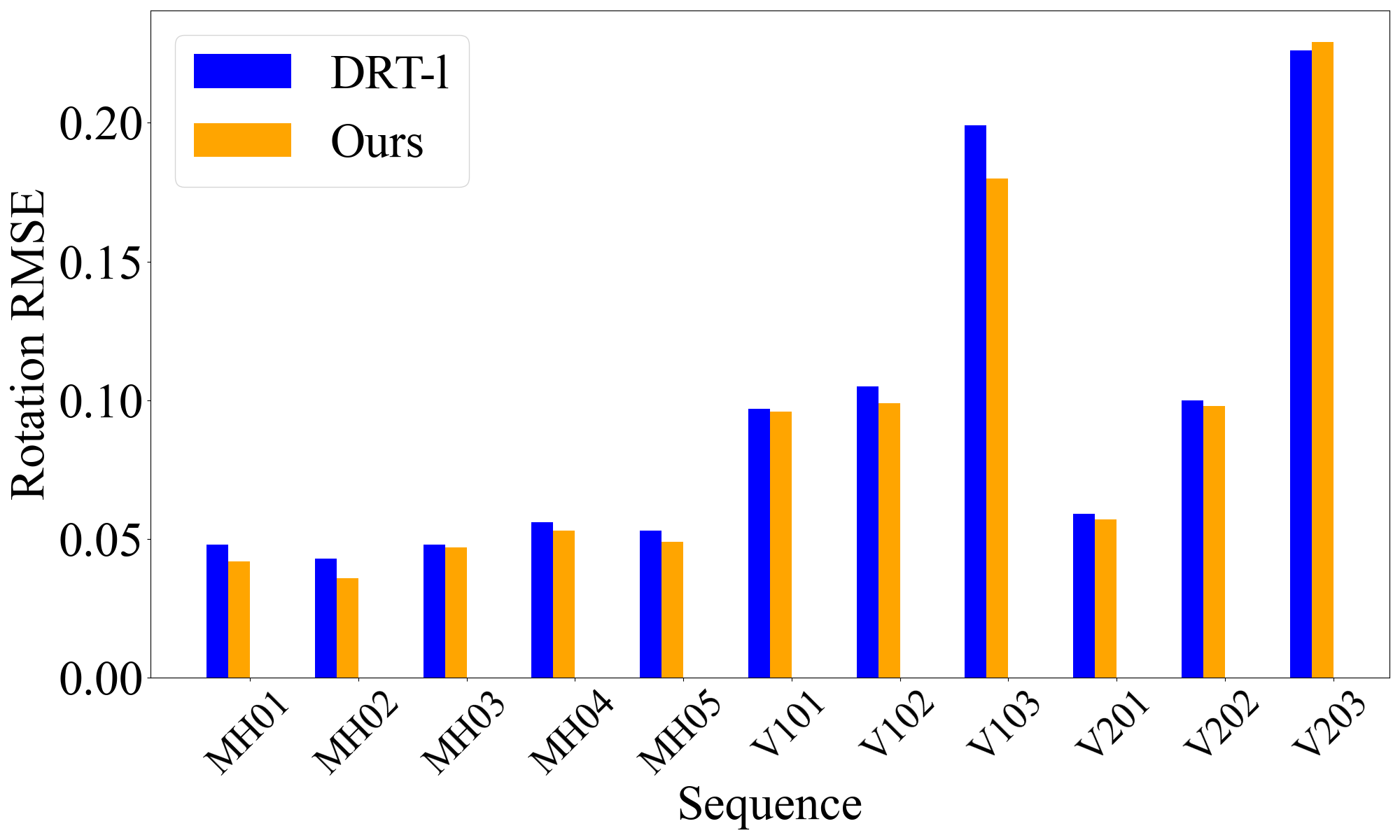}
    \end{minipage}\hfill   
    \caption{Gyroscope bias errors and rotation RMSE on EuRoC sequences.}
    \label{pic:2}
\end{figure*}

\subsection{Scale and Gravity Refinement}
Accurate gravity estimation is needed to improve VIO performance as it affects translation/velocity observability and integration. A precise scale factor is also required to align visual structure with metric scale, enhancing accuracy. Therefore, we introduce gravity magnitude $G$ \cite{mur2017visual} to refine both gravity and scale. Let $\mathbf{g}^{\mathrm{I}}=\{0,0,1\}$ be the gravity direction of the inertial reference $\mathrm{I}$. Based on the gravity $\mathbf{g}^{c_{0}}$ from previous step, we can calculate $\mathbf{R}_{c_{0}\mathrm{I}}$:
\begin{equation}
\begin{array}{c}
\mathbf{R}_{c_{0}\mathrm{I}}=\operatorname{Exp}(\mathbf{v}\theta) \\\\
\mathbf{v}=\frac{\mathbf{g}^{\mathrm{I}} \times \mathbf{g}^{c_{0}}}{\left\|\mathbf{g}^{\mathrm{I}} \times \mathbf{g}^{c_{0}}\right\|}, \quad \theta=\operatorname{atan} 2\left(\left\|\mathbf{g}^{\mathrm{I}} \times \mathbf{g}^{c_{0}}\right\|, \mathbf{g}^{\mathrm{I}} \cdot \mathbf{g}^{c_{0}}\right)
\end{array}
\end{equation}
the new gravity is then expressed as:
\begin{equation}
\hat{\mathbf{g}}^{c_{0}}=\mathbf{R}_{c_{0}\mathrm{I}} \mathbf{g}^{\mathrm{I}} G
\end{equation}

The rotation matrix $\mathbf{R}_{c_{0}\mathrm{I}}$ can be parametrized with just two angles around the x and y axes in $\mathrm{I}$, as a rotation around the z axis has no effect in $\mathbf{g}^{c_{0}}$ \cite{mur2017visual}. By introducing a perturbation $\boldsymbol{\delta} \boldsymbol{\theta}$, we can optimize the rotation as follows:
\begin{equation}
\begin{aligned}
\hat{\mathbf{g}}^{c_{0}}=\mathbf{R}_{c_{0}\mathrm{I}} \operatorname{Exp}(\boldsymbol{\delta} \boldsymbol{\theta}) \mathbf{g}^{\mathrm{I}} G \\
\boldsymbol{\delta} \boldsymbol{\theta}=\left[\begin{array}{ll}
\boldsymbol{\delta} \boldsymbol{\theta}_{\boldsymbol{x} \boldsymbol{y}}^{\top} & 0
\end{array}\right]^{\top} \\\boldsymbol{\delta} \boldsymbol{\theta}_{\boldsymbol{x} \boldsymbol{y}}=\left[\begin{array}{ll}
\delta \theta_{x} & \delta \theta_{y}
\end{array}\right]^{\top} \\
\end{aligned}
\end{equation}
with a first-order approximation:
\begin{equation}
\hat{\mathbf{g}}^{c_{0}} \approx \mathbf{R}_{c_{0}\mathrm{I}} \mathbf{g}^{\mathrm{I}} G-\mathbf{R}_{c_{0}\mathrm{I}}\left\lfloor\mathbf{g}^{\mathrm{I}}\right\rfloor_{\times} G \boldsymbol{\delta} \boldsymbol{\theta}
\label{eq:29}
\end{equation}

Substituting $s \mathbf{p}_{c_{0} b_{k}}=s \mathbf{p}_{c_{0} c_{k}}-\mathbf{R}_{c_{0} b_{k}} \mathbf{p}_{b c}$ into Eq. \eqref{eq:5},  we have:
\begin{equation}
\begin{split}
s\mathbf{p}_{c_{0} c_{j}}&=s\mathbf{p}_{c_{0} c_{i}}+\mathbf{v}_{b_{i}}^{c_{0}}\Delta t_{i j}-\frac{1}{2} \mathbf{g}^{c_{0}} \Delta t_{i j}^{2}\\&+\mathbf{R}_{c_{0} b_{i}} \boldsymbol{\alpha}_{b_{j}}^{b_{i}}+\left(\mathbf{R}_{c_{0} b_{j}}-\mathbf{R}_{c_{0} b_{i}}\right) \mathbf{p}_{b c}
\end{split}
\label{eq:30}
\end{equation}
Then according to Eqs. \eqref{eq:29} and \eqref{eq:30}, we can derive:
\begin{equation}
\begin{split}
s\mathbf{p}_{c_{0} c_{j}}&=s\mathbf{p}_{c_{0} c_{i}}+\mathbf{v}_{b_{i}}^{c_{0}}\Delta t_{i j}+\frac{1}{2} \mathbf{R}_{c_{0}\mathrm{I}}\left\lfloor\mathbf{g}^{\mathrm{I}}\right\rfloor_{\times} G \Delta t_{i j}^{2}\boldsymbol{\delta} \boldsymbol{\theta}\\&-\frac{1}{2}\mathbf{R}_{c_{0}\mathrm{I}} \mathbf{g}^{\mathrm{I}} G\Delta t_{i j}^{2}+\mathbf{R}_{c_{0} b_{i}} \boldsymbol{\alpha}_{b_{j}}^{b_{i}}+\left(\mathbf{R}_{c_{0} b_{j}}{-}\mathbf{R}_{c_{0} b_{i}}\right) \mathbf{p}_{b c}
\end{split}
\end{equation}

We consider two connections between three consecutive keyframes $i$, $j$, and $k$. Using Eq. \eqref{eq:6}, we eliminate the velocity, resulting in the following equation:
\begin{equation}
\left[\begin{array}{ll}
\boldsymbol{\lambda}(i) & \boldsymbol{\phi}(i)
\end{array}\right]\left[\begin{array}{c}
s \\
\boldsymbol{\delta} \boldsymbol{\theta}_{\boldsymbol{x} \boldsymbol{y}}
\end{array}\right]=\boldsymbol{\psi}(i)
\end{equation}
where
\begin{equation}
\begin{aligned}
\boldsymbol{\lambda}(i) & =\left(\mathbf{p}_{c_{0} c_{j}}-\mathbf{p}_{c_{0} c_{i}}\right) \Delta t_{jk}-\left(\mathbf{p}_{c_{0} c_{k}}-\mathbf{p}_{c_{0} c_{j}}\right) \Delta t_{ij} \\
\boldsymbol{\phi}(i) & =\frac{1}{2} \mathbf{R}_{c_{0}\mathrm{I}}\left\lfloor\mathbf{g}^{\mathrm{I}}\right\rfloor_{\times} G\left(\Delta t_{ij}^{2} \Delta t_{jk}+\Delta t_{jk}^{2} \Delta t_{ij}\right) \\
\boldsymbol{\psi}(i) & =\left(\mathbf{R}_{c_{0} b_{j}}{-}\mathbf{R}_{c_{0} b_{i}}\right) \mathbf{p}_{b c} \Delta t_{jk}{-}\left(\mathbf{R}_{c_{0} b_{k}}{-}\mathbf{R}_{c_{0} b_{j}}\right)\mathbf{p}_{b c} \Delta t_{ij} \\
& +\mathbf{R}_{c_{0} b_{i}} \boldsymbol{\alpha}_{b_{j}}^{b_{i}} \Delta t_{jk}{-}\mathbf{R}_{c_{0} b_{i}} \boldsymbol{\beta}_{b_{j}}^{b_{i}} \Delta t_{ij} \Delta t_{jk} 
{-}\mathbf{R}_{c_{0} b_{j}} \boldsymbol{\alpha}_{b_{k}}^{b_{j}} \Delta t_{ij}\\&-\frac{1}{2}\mathbf{R}_{c_{0}\mathrm{I}} \mathbf{g}^{\mathrm{I}} G\left(\Delta t_{ij}^{2} \Delta t_{jk}+\Delta t_{jk}^{2} \Delta t_{ij}\right)
\end{aligned}
\end{equation}

In initialization frames, each three consecutive keyframes form an equation. Combining them gives a system of linear equations, which can be solved by SVD to determine scale factor $s$ and gravity direction correction $\boldsymbol{\delta} \boldsymbol{\theta}_{\boldsymbol{x} \boldsymbol{y}}$. Finally, updating $\mathbf{R}_{c_{0}\mathrm{I}}$ completes scale and gravity refinement.

\begin{table*}[h]
\renewcommand\arraystretch{1.1}
\centering
\resizebox{\textwidth}{!}{%
\begin{tabular}{ccccccccccccc}
\hline
{\color[HTML]{333333} } &
  {\color[HTML]{333333} \textbf{Dataset}} &
  {\color[HTML]{333333} MH01} &
  {\color[HTML]{333333} MH02} &
  {\color[HTML]{333333} MH03} &
  {\color[HTML]{333333} MH04} &
  {\color[HTML]{333333} MH05} &
  {\color[HTML]{333333} V101} &
  {\color[HTML]{333333} V102} &
  {\color[HTML]{333333} V103} &
  {\color[HTML]{333333} V201} &
  {\color[HTML]{333333} V202} &
  {\color[HTML]{333333} V203} \\ \hline
\multicolumn{1}{c|}{{\color[HTML]{333333} }} &
\multicolumn{1}{c|}{\cellcolor[HTML]{FFFFFF}{\color[HTML]{333333} VINS-Mono}} &
  {\color[HTML]{333333} 0.147} &
  {\color[HTML]{333333} 0.165} &
  {\color[HTML]{333333} 0.166} &
  {\color[HTML]{333333} 0.169} &
  {\color[HTML]{333333} 0.195} &
  {\color[HTML]{333333} 0.143} &
  {\color[HTML]{333333} 0.150} &
  {\color[HTML]{333333} 0.252} &
  {\color[HTML]{333333} 0.196} &
  {\color[HTML]{333333} 0.174} &
  {\color[HTML]{333333} 0.281}  \\
\multicolumn{1}{c|}{{\color[HTML]{333333} }} &
  \multicolumn{1}{c|}{\cellcolor[HTML]{FFFFFF}{\color[HTML]{333333} DRT-t}} &
  {\color[HTML]{333333} 0.188} &
  {\color[HTML]{333333} 0.158} &
  {\color[HTML]{333333} 0.121} &
  {\color[HTML]{333333} 0.226} &
  {\color[HTML]{333333} 0.240} &
  {\color[HTML]{333333} 0.149} &
  {\color[HTML]{FF0000} \textbf{0.061}} &
  {\color[HTML]{FF0000} \textbf{0.113}} &
  {\color[HTML]{333333} 0.146} &
  {\color[HTML]{FF0000} \textbf{0.065}} &
  {\color[HTML]{FF0000} \textbf{0.106}} \\
\multicolumn{1}{c|}{{\color[HTML]{333333} }} &
  \multicolumn{1}{c|}{\cellcolor[HTML]{FFFFFF}{\color[HTML]{333333} DRT-l}} &
  \cellcolor[HTML]{FFFFFF}{\color[HTML]{0002FF} 0.097} &
  \cellcolor[HTML]{FFFFFF}{\color[HTML]{0002FF} 0.112} &
  \cellcolor[HTML]{FFFFFF}{\color[HTML]{0002FF} 0.085} &
  \cellcolor[HTML]{FFFFFF}{\color[HTML]{0002FF} 0.167} &
  \cellcolor[HTML]{FFFFFF}{\color[HTML]{0002FF} 0.164} &
  \cellcolor[HTML]{FFFFFF}{\color[HTML]{0002FF} 0.113} &
  \cellcolor[HTML]{FFFFFF}{\color[HTML]{0002FF} 0.078} &
  \cellcolor[HTML]{FFFFFF}{\color[HTML]{0002FF} 0.178} &
  \cellcolor[HTML]{FFFFFF}{\color[HTML]{0002FF} 0.106} &
  \cellcolor[HTML]{FFFFFF}{\color[HTML]{333333} 0.085} &
  \cellcolor[HTML]{FFFFFF}{\color[HTML]{0002FF} 0.156} \\
\multicolumn{1}{c|}{\multirow{-4}{*}{{\color[HTML]{333333} \textbf{\begin{tabular}[c]{@{}c@{}}Scale\\ RMSE\end{tabular}}}}} &
  \multicolumn{1}{c|}{\cellcolor[HTML]{FFFFFF}{\color[HTML]{333333} \textbf{Ours}}} &
  \cellcolor[HTML]{FFFFFF}{\color[HTML]{FF0000} \textbf{0.095}} &
  \cellcolor[HTML]{FFFFFF}{\color[HTML]{FF0000} \textbf{0.094}} &
  \cellcolor[HTML]{FFFFFF}{\color[HTML]{FF0000} \textbf{0.081}}&
  \cellcolor[HTML]{FFFFFF}{\color[HTML]{FF0000} \textbf{0.165}} &
  \cellcolor[HTML]{FFFFFF}{\color[HTML]{FF0000} \textbf{0.157}} &
  \cellcolor[HTML]{FFFFFF}{\color[HTML]{FF0000} \textbf{0.111}} &
  \cellcolor[HTML]{FFFFFF}{\color[HTML]{333333} 0.080} &
  \cellcolor[HTML]{FFFFFF}{\color[HTML]{0002FF} 0.178} &
  \cellcolor[HTML]{FFFFFF}{\color[HTML]{FF0000} \textbf{0.105}} &
  \cellcolor[HTML]{FFFFFF}{\color[HTML]{0002FF} 0.081} &
  \cellcolor[HTML]{FFFFFF}{\color[HTML]{333333} 0.178} \\ \hline
\multicolumn{1}{c|}{{\color[HTML]{333333} }} &
\multicolumn{1}{c|}{\cellcolor[HTML]{FFFFFF}{\color[HTML]{333333} VINS-Mono}} &
  {\color[HTML]{333333} 0.063} &
  {\color[HTML]{333333} 0.071} &
  {\color[HTML]{333333} 0.175} &
  {\color[HTML]{333333} 0.156} &
  {\color[HTML]{333333} 0.140} &
  {\color[HTML]{333333} 0.062} &
  {\color[HTML]{333333} 0.145} &
  {\color[HTML]{333333} 0.211} &
  {\color[HTML]{333333} 0.061} &
  {\color[HTML]{333333} 0.075} &
  {\color[HTML]{333333} 0.142}  \\
\multicolumn{1}{c|}{{\color[HTML]{333333} }} &
  \multicolumn{1}{c|}{\cellcolor[HTML]{FFFFFF}{\color[HTML]{333333} DRT-t}} &
  {\color[HTML]{333333} 0.088} &
  {\color[HTML]{333333} 0.080} &
  {\color[HTML]{333333} 0.125} &
  {\color[HTML]{333333} 0.197} &
  {\color[HTML]{333333} 0.182} &
  {\color[HTML]{333333} 0.069} &
  {\color[HTML]{FF0000} \textbf{0.072}} &
  {\color[HTML]{FF0000} \textbf{0.128}} &
  {\color[HTML]{333333} 0.062} &
  {\color[HTML]{333333} 0.056} &
  {\color[HTML]{FF0000} \textbf{0.095}} \\
\multicolumn{1}{c|}{{\color[HTML]{333333} }} &
  \multicolumn{1}{c|}{\cellcolor[HTML]{FFFFFF}{\color[HTML]{333333} DRT-l}} &
  \cellcolor[HTML]{FFFFFF}{\color[HTML]{0002FF} 0.052} &
  \cellcolor[HTML]{FFFFFF}{\color[HTML]{0002FF} 0.056} &
  \cellcolor[HTML]{FFFFFF}{\color[HTML]{0002FF} 0.092} &
  \cellcolor[HTML]{FFFFFF}{\color[HTML]{0002FF} 0.155} &
  \cellcolor[HTML]{FFFFFF}{\color[HTML]{0002FF} 0.135} &
  \cellcolor[HTML]{FFFFFF}{\color[HTML]{0002FF} 0.050} &
  \cellcolor[HTML]{FFFFFF}{\color[HTML]{333333} 0.077} &
  \cellcolor[HTML]{FFFFFF}{\color[HTML]{333333} 0.150} &
  \cellcolor[HTML]{FFFFFF}{\color[HTML]{FF0000} \textbf{0.043}} &
  \cellcolor[HTML]{FFFFFF}{\color[HTML]{0002FF} 0.055} &
  \cellcolor[HTML]{FFFFFF}{\color[HTML]{0002FF} 0.101} \\
\multicolumn{1}{c|}{\multirow{-4}{*}{{\color[HTML]{333333} \textbf{\begin{tabular}[c]{@{}c@{}}Velocity\\ RMSE\\ $(m/s)$\end{tabular}}}}} &
  \multicolumn{1}{c|}{\cellcolor[HTML]{FFFFFF}{\color[HTML]{333333} \textbf{Ours}}} &
  \cellcolor[HTML]{FFFFFF}{\color[HTML]{FF0000} \textbf{0.051}} &
  \cellcolor[HTML]{FFFFFF}{\color[HTML]{FF0000} \textbf{0.051}} &
  \cellcolor[HTML]{FFFFFF}{\color[HTML]{FF0000} \textbf{0.087}} &
  \cellcolor[HTML]{FFFFFF}{\color[HTML]{FF0000} \textbf{0.154}} &
  \cellcolor[HTML]{FFFFFF}{\color[HTML]{FF0000} \textbf{0.133}} &
  \cellcolor[HTML]{FFFFFF}{\color[HTML]{FF0000} \textbf{0.048}} &
  \cellcolor[HTML]{FFFFFF}{\color[HTML]{0002FF} 0.076} &
  \cellcolor[HTML]{FFFFFF}{\color[HTML]{0002FF} 0.140} &
  \cellcolor[HTML]{FFFFFF}{\color[HTML]{0002FF} 0.044} &
  \cellcolor[HTML]{FFFFFF}{\color[HTML]{FF0000} \textbf{0.054}} &
  \cellcolor[HTML]{FFFFFF}{\color[HTML]{333333} 0.102} \\ \hline
\multicolumn{1}{c|}{{\color[HTML]{333333} }} &
\multicolumn{1}{c|}{\cellcolor[HTML]{FFFFFF}{\color[HTML]{333333} VINS-Mono}} &
  {\color[HTML]{333333} 1.172} &
  {\color[HTML]{333333} 1.112} &
  {\color[HTML]{333333} 1.486} &
  {\color[HTML]{333333} 1.206} &
  {\color[HTML]{333333} 1.311} &
  {\color[HTML]{333333} 3.205} &
  {\color[HTML]{333333} 2.544} &
  {\color[HTML]{333333} 2.688} &
  {\color[HTML]{333333} 1.358} &
  {\color[HTML]{333333} 1.258} &
  {\color[HTML]{333333} 4.355}  \\
\multicolumn{1}{c|}{{\color[HTML]{333333} }} &
  \multicolumn{1}{c|}{\cellcolor[HTML]{FFFFFF}{\color[HTML]{333333} DRT-t}} &
  {\color[HTML]{333333} 0.959} &
  {\color[HTML]{333333} 0.949} &
  {\color[HTML]{333333} 0.931} &
  {\color[HTML]{333333} 1.087} &
  {\color[HTML]{333333} 0.992} &
  {\color[HTML]{FF0000} \textbf{3.155}} &
  {\color[HTML]{0002FF} 0.850} &
  {\color[HTML]{0002FF} 1.657} &
  {\color[HTML]{333333} 1.064} &
  {\color[HTML]{FF0000} \textbf{0.856}} &
  {\color[HTML]{333333} 1.278} \\
\multicolumn{1}{c|}{{\color[HTML]{333333} }} &
  \multicolumn{1}{c|}{\cellcolor[HTML]{FFFFFF}{\color[HTML]{333333} DRT-l}} &
  \cellcolor[HTML]{FFFFFF}{\color[HTML]{0002FF} 0.938} &
  \cellcolor[HTML]{FFFFFF}{\color[HTML]{0002FF} 0.934} &
  \cellcolor[HTML]{FFFFFF}{\color[HTML]{0002FF} 0.863} &
  \cellcolor[HTML]{FFFFFF}{\color[HTML]{0002FF} 1.001}&
  \cellcolor[HTML]{FFFFFF}{\color[HTML]{0002FF} 0.901} &
  \cellcolor[HTML]{FFFFFF}{\color[HTML]{333333} 3.215} &
  \cellcolor[HTML]{FFFFFF}{\color[HTML]{333333} 0.861} &
  \cellcolor[HTML]{FFFFFF}{\color[HTML]{333333} 1.798} &
  \cellcolor[HTML]{FFFFFF}{\color[HTML]{0002FF} 1.052} &
  \cellcolor[HTML]{FFFFFF}{\color[HTML]{0002FF} 0.956} &
  \cellcolor[HTML]{FFFFFF}{\color[HTML]{FF0000} \textbf{1.065}} \\
\multicolumn{1}{c|}{\multirow{-4}{*}{{\color[HTML]{333333} \textbf{\begin{tabular}[c]{@{}c@{}}G.Dir\\ RMSE\\ (°)\end{tabular}}}}} &
  \multicolumn{1}{c|}{\cellcolor[HTML]{FFFFFF}{\color[HTML]{333333} \textbf{Ours}}} &
  \cellcolor[HTML]{FFFFFF}{\color[HTML]{FF0000} \textbf{0.652}} &
  \cellcolor[HTML]{FFFFFF}{\color[HTML]{FF0000} \textbf{0.621}} &
  \cellcolor[HTML]{FFFFFF}{\color[HTML]{FF0000} \textbf{0.606}} &
  \cellcolor[HTML]{FFFFFF}{\color[HTML]{FF0000} \textbf{0.704}} &
  \cellcolor[HTML]{FFFFFF}{\color[HTML]{FF0000} \textbf{0.630}} &
  \cellcolor[HTML]{FFFFFF}{\color[HTML]{FF0000} \textbf{2.752}} &
  \cellcolor[HTML]{FFFFFF}{\color[HTML]{FF0000} \textbf{0.727}} &
  \cellcolor[HTML]{FFFFFF}{\color[HTML]{FF0000} \textbf{1.399}} &
  \cellcolor[HTML]{FFFFFF}{\color[HTML]{FF0000} \textbf{0.808}} &
  \cellcolor[HTML]{FFFFFF}{\color[HTML]{333333} 1.014} &
  \cellcolor[HTML]{FFFFFF}{\color[HTML]{0002FF} 1.091} \\ \hline
\end{tabular}%
}
\caption{Detailed initialization results for the 10KFs setting in each dataset from EuRoC. For each metric, the best result is highlighted in \textcolor{red}{\textbf{red}}, the second best in \textcolor{blue}{blue}.}
\label{tab:1}
\end{table*}

\begin{table*}[h]
\centering
\resizebox{0.9\textwidth}{!}{%
\begin{tabular}{ccccccccccccc}
\hline
\textbf{} &
  \multicolumn{3}{c}{G.Dir RMSE} &
  \multicolumn{3}{c}{Pose RMSE} &
  \multicolumn{3}{c}{Rotation RMSE} &
  \multicolumn{3}{c}{Scale RMSE} \\
 &
  DRT-l &
  DRT-t &
  \textbf{Ours} &
  DRT-l &
  DRT-t &
  \textbf{Ours} &
  DRT-l &
  DRT-t &
  \textbf{Ours} &
  DRT-l &
  DRT-t &
  \textbf{Ours} \\ \hline
\multicolumn{1}{c|}{room1} &
  0.928 &
  {\color[HTML]{FF0000} \textbf{0.597}} &
  \multicolumn{1}{c|}{0.758} &
  0.039 &
  {\color[HTML]{FF0000} \textbf{0.022}} &
  \multicolumn{1}{c|}{0.031} &
  0.289 &
  0.264 &
  \multicolumn{1}{c|}{{\color[HTML]{FF0000} \textbf{0.255}}} &
  0.065 &
  {\color[HTML]{FF0000} \textbf{0.038}} &
  0.057 \\
\multicolumn{1}{c|}{room2} &
  0.896 &
  1.136 &
  \multicolumn{1}{c|}{{\color[HTML]{FF0000} \textbf{0.811}}} &
  0.046 &
  0.062 &
  \multicolumn{1}{c|}{{\color[HTML]{FF0000} \textbf{0.044}}} &
  0.371 &
  {\color[HTML]{FF0000} \textbf{0.321}} &
  \multicolumn{1}{c|}{{\color[HTML]{FF0000} \textbf{0.321}}} &
  {\color[HTML]{FF0000} \textbf{0.089}} &
  0.111 &
  0.104 \\
\multicolumn{1}{c|}{room3} &
  1.045 &
  1.489 &
  \multicolumn{1}{c|}{{\color[HTML]{FF0000} \textbf{0.878}}} &
  0.089 &
  0.109 &
  \multicolumn{1}{c|}{{\color[HTML]{FF0000} \textbf{0.085}}} &
  0.393 &
  0.381 &
  \multicolumn{1}{c|}{{\color[HTML]{FF0000} \textbf{0.368}}} &
  0.152 &
  0.176 &
  {\color[HTML]{FF0000} \textbf{0.148}} \\
\multicolumn{1}{c|}{room4} &
  0.935 &
  0.971 &
  \multicolumn{1}{c|}{{\color[HTML]{FF0000} \textbf{0.819}}} &
  0.056 &
  {\color[HTML]{FF0000} \textbf{0.046}} &
  \multicolumn{1}{c|}{{\color[HTML]{FF0000} \textbf{0.046}}} &
  0.279 &
  0.257 &
  \multicolumn{1}{c|}{{\color[HTML]{FF0000} \textbf{0.254}}} &
  0.145 &
  {\color[HTML]{FF0000} \textbf{0.105}} &
  0.129 \\
\multicolumn{1}{c|}{room5} &
  0.952 &
  {\color[HTML]{FF0000} \textbf{0.858}} &
  \multicolumn{1}{c|}{0.893} &
  0.039 &
  0.034 &
  \multicolumn{1}{c|}{{\color[HTML]{FF0000} \textbf{0.033}}} &
  0.396 &
  0.367 &
  \multicolumn{1}{c|}{{\color[HTML]{FF0000} \textbf{0.365}}} &
  0.094 &
  0.081 &
  {\color[HTML]{FF0000} \textbf{0.078}} \\
\multicolumn{1}{c|}{room6} &
  0.595 &
  0.447 &
  \multicolumn{1}{c|}{{\color[HTML]{FF0000} \textbf{0.408}}} &
  0.031 &
  0.027 &
  \multicolumn{1}{c|}{{\color[HTML]{FF0000} \textbf{0.026}}} &
  0.298 &
  {\color[HTML]{FF0000} \textbf{0.241}} &
  \multicolumn{1}{c|}{0.258} &
  0.105 &
  0.093 &
  {\color[HTML]{FF0000} \textbf{0.091}} \\
\multicolumn{1}{c|}{corridor1} &
  1.039 &
  1.626 &
  \multicolumn{1}{c|}{{\color[HTML]{FF0000} \textbf{0.966}}} &
  {\color[HTML]{FF0000} \textbf{0.033}} &
  0.041 &
  \multicolumn{1}{c|}{{\color[HTML]{FF0000} \textbf{0.033}}} &
  0.338 &
  0.343 &
  \multicolumn{1}{c|}{{\color[HTML]{FF0000} \textbf{0.328}}} &
  0.089 &
  0.075 &
  {\color[HTML]{FF0000} \textbf{0.073}} \\
\multicolumn{1}{c|}{corridor2} &
  1.089 &
  0.869 &
  \multicolumn{1}{c|}{{\color[HTML]{FF0000} \textbf{0.952}}} &
  0.045 &
  {\color[HTML]{FF0000} \textbf{0.034}} &
  \multicolumn{1}{c|}{0.037} &
  0.364 &
  0.366 &
  \multicolumn{1}{c|}{{\color[HTML]{FF0000} \textbf{0.355}}} &
  0.111 &
  {\color[HTML]{FF0000} \textbf{0.069}} &
  0.091 \\
\multicolumn{1}{c|}{corridor3} &
  0.917 &
  {\color[HTML]{FF0000} \textbf{0.848}} &
  \multicolumn{1}{c|}{0.867} &
  0.051 &
  {\color[HTML]{FF0000} \textbf{0.045}} &
  \multicolumn{1}{c|}{0.052} &
  0.401 &
  0.405 &
  \multicolumn{1}{c|}{{\color[HTML]{FF0000} \textbf{0.388}}} &
  0.099 &
  {\color[HTML]{FF0000} \textbf{0.095}} &
  0.113 \\
\multicolumn{1}{c|}{corridor4} &
  1.028 &
  {\color[HTML]{FF0000} \textbf{0.809}} &
  \multicolumn{1}{c|}{0.821} &
  0.047 &
  0.035 &
  \multicolumn{1}{c|}{{\color[HTML]{FF0000} \textbf{0.032}}} &
  0.322 &
  0.299 &
  \multicolumn{1}{c|}{{\color[HTML]{FF0000} \textbf{0.297}}} &
  0.131 &
  {\color[HTML]{FF0000} \textbf{0.091}} &
  0.103 \\
\multicolumn{1}{c|}{corridor5} &
  0.683 &
  0.738 &
  \multicolumn{1}{c|}{{\color[HTML]{FF0000} \textbf{0.501}}} &
  0.031 &
  {\color[HTML]{FF0000} \textbf{0.026}} &
  \multicolumn{1}{c|}{0.035} &
  0.277 &
  0.267 &
  \multicolumn{1}{c|}{{\color[HTML]{FF0000} \textbf{0.261}}} &
  0.085 &
  {\color[HTML]{FF0000} \textbf{0.071}} &
  0.115 \\ \hline
\end{tabular}%
}
\caption{Initialization results for the 10KFs setting in TUM Visual-Inertial dataset. The best result is highlighted in \textcolor{red}{\textbf{red}}.}
\label{tab:5}
\end{table*}

\section{Experiments}
In this section, we evaluate our IMU initialization method on the EuRoC dataset \cite{burri2016euroc} and the TUM VI dataset \cite{schubert2018tum}. Both datasets provide camera images at 20Hz, IMU data at 200Hz, and ground-truth trajectories. The EuRoC dataset contains 11 sequences. In order to verify the generalization ability, we also select 11 sequences from the TUM dataset that are in different scenarios. We divide them into segments with 10 keyframes sampled at 4Hz \cite{he2023rotation}. We evaluate performance using gyroscope bias, velocity, gravity, and scale errors. Scale error is computed with Umeyama alignment \cite{umeyama1991least}. An initialization is considered successful when the scale error is less than one, and the Root Mean Square Error (RMSE) is employed to evaluate the method. We compare with DRT-t, DRT-l in \cite{he2023rotation}, and VINS-Mono \cite{qin2018vins}. All algorithms use the same image processing. We track existing features via the KLT sparse optical flow algorithm \cite{lucas1981iterative} and detect new corner features \cite{shi1994good} to keep 150 points per image. RANSAC with a fundamental matrix model \cite{hartley2003multiple} is used to reduce outliers. All experiments run on an Intel i7-9700 desktop with 32 GB of RAM. 

\begin{table*}[h]
\centering
\resizebox{0.7\textwidth}{!}{%
\begin{tabular}{ccllcllcccccc}
\hline
\textbf{}                  & \multicolumn{3}{c}{Bg Est.}                   & \multicolumn{3}{c}{Sca\&Grav Ref.} & Bg    & Rotation & Scale & Velocity & G.Dir & SUM   \\ \hline
\multicolumn{1}{c|}{DRT-t} & \multicolumn{3}{c}{-}                         & \multicolumn{3}{c|}{-}             & 2.192 & 0.099    & 0.143 & 0.105    & 1.252 & 3.791 \\
\multicolumn{1}{c|}{DRT-t} & \multicolumn{3}{c}{\checkmark} & \multicolumn{3}{c|}{-}             & 2.115 & 0.097    & 0.144 & 0.103    & 1.236 & 3.695 \\
\multicolumn{1}{c|}{DRT-l} & \multicolumn{3}{c}{-}                         & \multicolumn{3}{c|}{-}             & 1.881 & 0.091    & 0.121 & 0.088    & 1.235 & 3.416 \\
\multicolumn{1}{c|}{DRT-l}  & \multicolumn{3}{c}{\checkmark} & \multicolumn{3}{c|}{-}             & {\color[HTML]{FF0000} \textbf{1.724}} & {\color[HTML]{FF0000} \textbf{0.087}}    & 0.120 & 0.091    & 1.249 & 3.271 \\
\multicolumn{1}{c|}{DRT-l} &
  \multicolumn{3}{c}{-} &
  \multicolumn{3}{c|}{\checkmark} &
  1.881 &
  0.091 &
  0.119 &
  {\color[HTML]{FF0000} \textbf{0.086}} &
  {\color[HTML]{FF0000} \textbf{1.001}} &
  3.178 \\
\multicolumn{1}{c|}{\textbf{Ours}} &
  \multicolumn{3}{c}{\checkmark} &
  \multicolumn{3}{c|}{\checkmark} &
  {\color[HTML]{FF0000} \textbf{1.724}} &
  {\color[HTML]{FF0000} \textbf{0.087}} &
  {\color[HTML]{FF0000} \textbf{0.118}} &
  0.090 &
  1.028 &
  {\color[HTML]{FF0000} \textbf{3.047}} \\ \hline
\end{tabular}%
}
\caption{Ablation Experiment was conducted on 11 sequences in the EuRoC dataset. Bg Est and Sca\&Grav Ref represent the two modules proposed in this paper. For each method, the average value of each metric across the 11 sequences was calculated. Sca\&Grav Ref is not applicable to DRT-t. For each metric, the best result is highlighted in \textcolor{red}{\textbf{red}}.}
\label{tab:4}
\end{table*}

\begin{table}[h]
\begin{tabular}{ccccccc}
\hline
IQR   & MH01 & MH02 & MH03 & MH04 & MH05 & \textbf{}                             \\ \hline
DRT-t & 0.198         & 0.156                                 & 0.110         & 0.279         & 0.313         &                                       \\
DRT-l & 0.096         & 0.095                                 & 0.080         & 0.125         & 0.182         &                                       \\
\textbf{Ours} &
  {\color[HTML]{FF0000} \textbf{0.094}} &
  {\color[HTML]{FF0000} \textbf{0.093}} &
  {\color[HTML]{FF0000} \textbf{0.078}} &
  {\color[HTML]{FF0000} \textbf{0.122}} &
  {\color[HTML]{FF0000} \textbf{0.175}} &
  {\color[HTML]{FF0000} \textbf{}} \\ \hline
IQR   & V101 & V102 & V103 & V201 & V202 & V203                     \\ \hline
DRT-t & 0.143         & {\color[HTML]{FF0000} \textbf{0.048}} & 0.158         & 0.118         & 0.067         & {\color[HTML]{FF0000} \textbf{0.104}} \\
DRT-l & 0.120         & 0.059                                 & 0.288         & 0.097         & 0.051         & 0.127                                 \\
\textbf{Ours} &
  {\color[HTML]{FF0000} \textbf{0.117}} &
  0.051 &
  {\color[HTML]{FF0000} \textbf{0.112}} &
  {\color[HTML]{FF0000} \textbf{0.095}} &
  {\color[HTML]{FF0000} \textbf{0.048}} &
  0.145 \\ \hline
\end{tabular}
\caption{The interquartile range (IQR) of scale RMSE of the successfully initialized segments on each sequence. The best result is highlighted in \textcolor{red}{\textbf{red}}.}
\label{tab:3}
\end{table}

\begin{table}[h]
\centering
\resizebox{\columnwidth}{!}{%
\begin{tabular}{ccccc}
\hline
Module         & VINS-Mono & DRT-t & DRT-l & \textbf{Ours} \\ \hline
SfM            & 29.24     & -     & -     & -    \\
3D Cov Gen.    & -         & -     & -     & 0.22 \\
Bg Est.        & 0.93      & 3.63  & 3.68  & 3.62 \\
Vel\&Grav Est. & 0.19      & 2.73  & 1.12  & 1.24 \\
Sca\&Grav Ref. &           & -     & -     & 0.15 \\ \hline
Total runtime  & 30.36     & 6.36  & 4.80   & 5.23 \\ \hline
\end{tabular}%
}
\caption{The average initialization time in milliseconds on the EuRoC dataset. We calculate the runtime for SfM, 3D covariance generation, gyroscope bias estimation, velocity and gravity estimation, and scale-gravity refinement.}
\label{tab:2}
\end{table}

\subsection{EuRoC dataset}
We compared the scale, gravity, and velocity estimated by the four methods on 1262 data segments in 11 sequences. From Table \ref{tab:1}, we can see that our method outperforms state-of-the-art initialization methods in almost all sequences, which verifies the effectiveness of our method. In terms of velocity and scale estimation, the tightly coupled method DRT-t is not as accurate as the loosely coupled DRT-l and ours. It is because DRT-t needs to integrate the IMU data from the initial moment to obtain velocity and position. This process is affected by noise, which degrades the accuracy of scale and velocity. In terms of gravity estimation, our method greatly reduces the gravity error and validates the effectiveness of the scale-gravity refinement module. We compare with DRT-l to verify the accuracy and robustness of our PNEC-based gyroscope bias estimation algorithm. Fig. \ref{pic:2} shows that in almost all sequences our method is more accurate than the previous best method DRT-l in gyroscope bias estimation. Owing to more precise gyroscope bias estimation, our method gets more accurate rotation via IMU pre-integration, which improves trajectory accuracy and VIO system performance.

\begin{figure*}[h]
    \centering
    \begin{tabular}{ccc}
        \hspace{-0.05\textwidth} \includegraphics[width=0.30\textwidth]{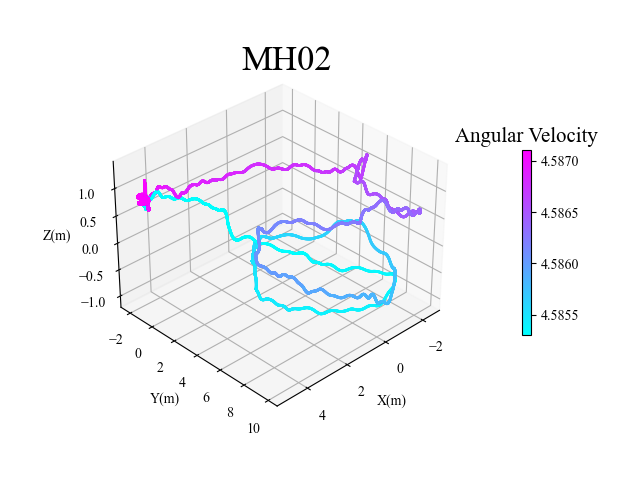} \hspace{-0.05\textwidth} &
        \includegraphics[width=0.30\textwidth]{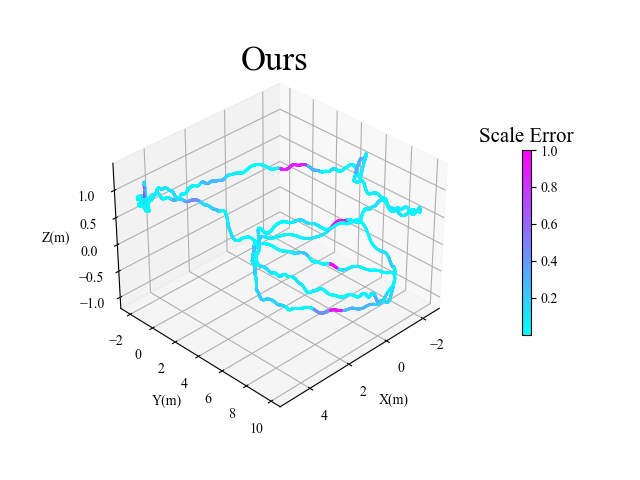} \hspace{-0.05\textwidth} &
        \includegraphics[width=0.30\textwidth]{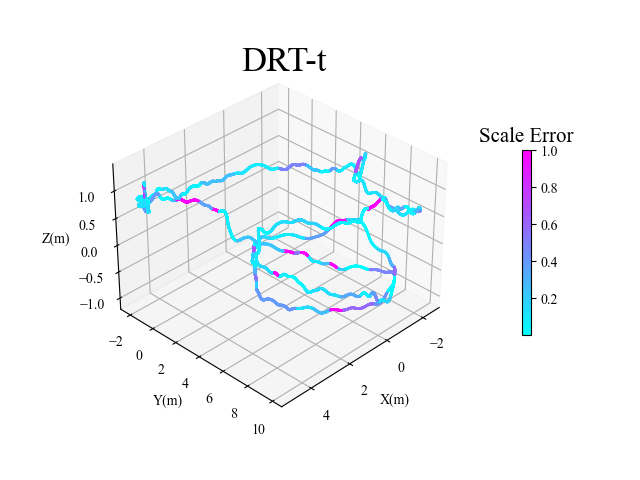} \vspace{-1.2em}
         \\ 
        \hspace{-0.04\textwidth} \includegraphics[width=0.30\textwidth]{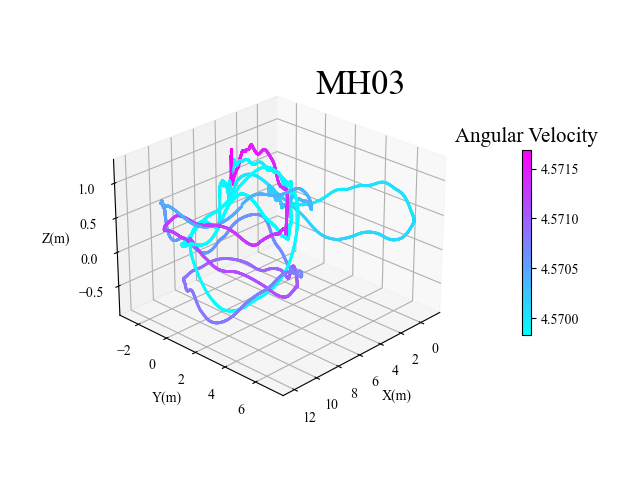} \hspace{-0.04\textwidth} &
        \includegraphics[width=0.30\textwidth]{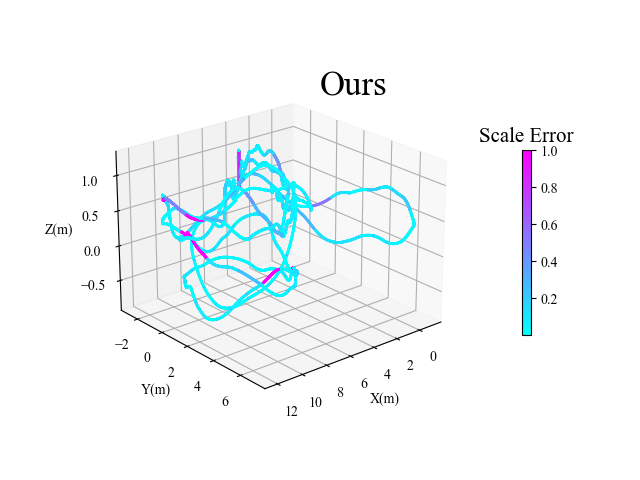} \hspace{-0.05\textwidth} &
        \includegraphics[width=0.30\textwidth]{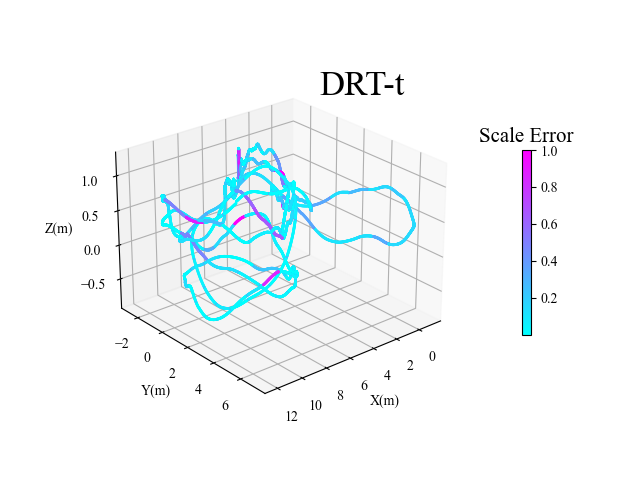} \vspace{-0.8em}
        \\
        (a) & (b) & (c)
    \end{tabular}
    \caption{Angular velocity and scale error visualizations for the MH02 dataset and MH03 dataset. The first-row image is MH02, and the second-row image is MH03. The column (a) shows the trajectory of the corresponding dataset colored based on the angular velocity. The columns (b) and (c) show the trajectory of our method and the DRT-t method based on the scale error colored on the corresponding dataset, respectively. The scale error is between 0 and 1. The lighter the color, the smaller the error.}
    \label{pic:3}
\end{figure*}

\subsection{TUM Visual-Inertial Dataset}
To evaluate the generalization ability of our method, we compare our method with DRT-t and DRT-l on the TUM VI dataset. As in Table \ref{tab:5}, our method achieves the lowest rotation error on most sequences, showing the efficacy of our gyroscope bias estimator. It also benefits pose estimation. For scale and gravity errors, DRT-t and our method each win half, similar to the results on EuRoC. We consider that since the DRT-t directly uses rotation and IMU pre-integration to solve for gravity direction without deriving translation, it may have more advantages in the case of large translation errors. Compared with DRT-l, our scale-gravity refinement module brings a significant improvement to the system performance, with the gravity direction error and scale error reduced by 14.2\% and 5.7\% on average, respectively. 

\subsection{Robustness Evaluation}
For robustness analysis, we visualize dataset trajectories and color them by scale errors from the solution. Scale error is key for evaluating initialization. We color sequences with errors under one and mark failures (purple) for errors $\ge $ 1. As in Fig. \ref{pic:2}, our algorithm has low errors and a high success rate across motions, even at high angular velocities. Compared to the tightly coupled DRT-t method, ours shows better robustness and accuracy in scale estimation. For quantitative comparison, we collect the scale errors of the successfully initialized data segments of each sequence and calculate the interquartile range of the scale errors. The interquartile range can describe the degree of dispersion in the middle part of the data and evaluate the robustness of all segments in each sequence. As shown in Table \ref{tab:3}, our method can estimate the scale more stably.

\subsection{Running Time Evaluation}
To demonstrate the runtime details of our method compared to DRT-l, DRT-t, and the initialization method of VINS-Mono, we separately calculate and sum the runtime of each module for comparison. Table \ref{tab:2} shows the runtime of each module in the 10KFs setup for four initialization methods. We can see that the initialization speed of DRT-l is still the fastest. Due to additional modules, our method is on average 0.43 milliseconds slower than DRT-l, but 1.13 milliseconds faster than DRT-t. This is because DRT-t requires long-time integration of accelerometer data from the initial moment. In conclusion, our initialization method meets the real-time performance requirements of VIO systems.

\subsection{Ablation Experiment}
In order to better evaluate the impact of the two proposed modules on systematic performance, we conduct ablation experiments on the EuRoC dataset. According to Table \ref{tab:4}, when DRT-t and DRT-l utilize the gyroscope bias estimator with 3D covariance, the gyroscope bias error and rotation error are reduced. The better rotation estimation leads to a lower velocity error and a lower gravity error in DRT-t, yet this does not apply to DRT-l. Nevertheless, our proposed scale-gravity refinement module successfully compensates for this and further reduces the errors. Our method combines the two proposed modules, successfully reducing the average error.

\section{Conclusion}
We propose a robust and accurate visual-inertial initialization method under a rotation-translation-decoupled framework. A gyroscope bias estimator with the Probabilistic Normal Epipolar Constraint (PNEC) is proposed, and a modified scale-gravity refinement module is incorporated. Benefiting from the new gyroscope bias estimator and the scale-gravity refinement module, which is also different from DRT-l, our method improves the accuracy and robustness while maintaining high computational efficiency. The experimental results on the EuRoC dataset and the TUM dataset prove that our method reduces the gyroscope bias error and the rotation error, thus reducing the velocity and pose error. The scale-gravity refinement module can significantly reduce gravity and scale error.



\bibliographystyle{./IEEEtran} 
\bibliography{paper}

\end{document}